\pdfoutput=1

\documentclass[11pt]{article}

\usepackage[final]{acl}

\usepackage{times}
\usepackage{latexsym}

\usepackage[T1]{fontenc}

\usepackage[utf8]{inputenc}

\usepackage{microtype}

\usepackage{inconsolata}

\usepackage{graphicx}

\usepackage{epsfig}
\usepackage{graphicx}
\usepackage{amsmath,bm}
\usepackage{amssymb}
\usepackage{multirow}
\usepackage{verbatim}
\usepackage{enumitem}
\usepackage{booktabs,amsfonts,dcolumn}
\newcommand\blfootnote[1]{%
  \begingroup
  \renewcommand\thefootnote{}\footnote{#1}%
  \addtocounter{footnote}{-1}%
  \endgroup
}

\usepackage{xspace}

\makeatletter
\DeclareRobustCommand\onedot{\futurelet\@let@token\@onedot}
\def\@onedot{\ifx\@let@token.\else.\null\fi\xspace}

\def\eg{\emph{e.g}\onedot} 
\def\ie{\emph{i.e}\onedot}

\makeatother

\usepackage{amssymb}
\usepackage{pifont}
\newcommand{\cmark}{\ding{51}}%
\newcommand{\xmark}{\ding{55}}%
\usepackage{multirow,multicol, makecell, booktabs}
\usepackage{array}
\usepackage{tcolorbox}

\usepackage{longtable}

\usepackage[capitalize]{cleveref}
\crefname{section}{Sec.}{Secs.}
\Crefname{section}{Section}{Sections}
\Crefname{table}{Table}{Tables}
\crefname{table}{Tab.}{Tabs.}

\usepackage{subcaption}

\title{Enhancing Temporal Modeling of Video LLMs via Time Gating}

\author{
Zi-Yuan Hu,
Yiwu Zhong,
Shijia Huang,
Michael R. Lyu, 
Liwei Wang$^{*}$ \\
Department of Computer Science and Engineering, The Chinese University of Hong Kong  \\
\texttt{\{zyhu22,sjhuang,lyu,lwwang\}@cse.cuhk.edu.hk}  \quad \texttt{yiwuzhong@cuhk.edu.hk}
}

\begin{document}
\maketitle
\begin{abstract}
Video Large Language Models (Video LLMs) have achieved impressive performance on video-and-language tasks, such as video question answering.
However, most existing Video LLMs neglect temporal information in video data, leading to struggles with temporal-aware video understanding.
To address this gap, we propose a \textbf{T}ime \textbf{G}ating \textbf{Vid}eo LLM (\textbf{TG-Vid}) designed to enhance temporal modeling through a novel \textbf{T}ime \textbf{G}ating module (\textbf{TG}).
The TG module employs a time gating mechanism on its sub-modules, comprising gating spatial attention, gating temporal attention, and gating MLP. This architecture enables our model to achieve a robust understanding of temporal information within videos.
Extensive evaluation of temporal-sensitive video benchmarks (\ie, MVBench, TempCompass, and NExT-QA) demonstrates that our TG-Vid model significantly outperforms the existing Video LLMs.
Further, comprehensive ablation studies validate that the performance gains are attributed to the designs of our TG module. 
Our code is available at \href{https://github.com/LaVi-Lab/TG-Vid}{https://github.com/LaVi-Lab/TG-Vid}.
\blfootnote{$^*$Corresponding author.}

\end{abstract}

\begin{figure*}[t]
    \centering
    \includegraphics[width=\linewidth]{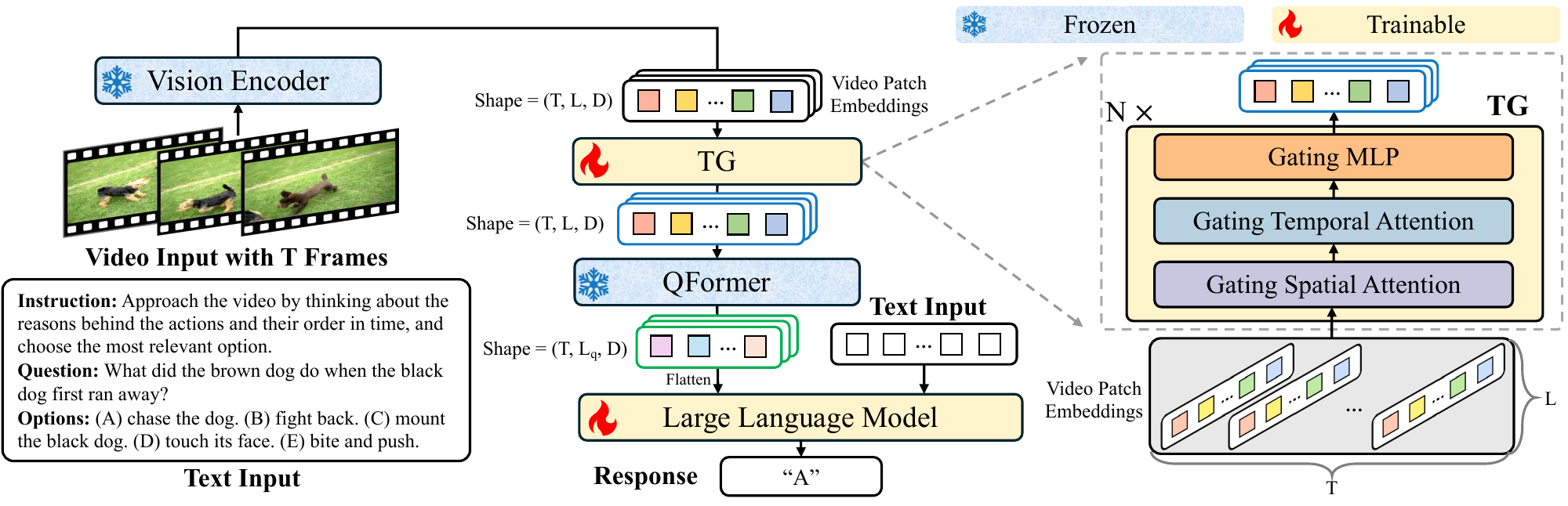}
    \caption{\textbf{Model architecture of TG-Vid}. 
    Given a video with $T$ frames, the vision encoder extracts $T$ frame-level embeddings.  
    Our TG employs a novel time gating mechanism to enhance video temporal modeling, thereby enhancing the frame-level video modeling ability of the QFormer.
    Moving forward, the QFormer compresses each frame-level video embedding from $L$ patch tokens to $L_\text{q}$ query tokens, followed by LLMs.
    }
    \label{fig:overview}
\end{figure*}

\section{Introduction}

The advancement of Large Language Models (LLMs)~\cite{llama,vicuna2023} has greatly fueled multi-modal research, such as Image LLMs~\cite{liu2024llava,bai2023qwen,InstructBLIP,liu2024improved} which have achieved success on image-and-language downstream tasks~\cite{goyal2017vqav2}.
Inspired by Image LLMs, many recent efforts manage to empower LLMs to understand video data~\cite{maaz2023videochatgpt,mvbench,stllm}. 
The typical architecture of these Video LLMs comprises a pretrained vision encoder~\cite{2021clip,evaclip}, a pretrained LLM~\cite{vicuna2023}, and a connection module in between~\cite{zhu2023minigpt4,InstructBLIP}.

Despite the impressive performance demonstrated by Video LLMs~\cite{maaz2023videochatgpt,xu2017MSRVTT_MSVD,yu2019activitynet,jang2017tgif}, a recent study~\cite{tempcompass} reveals that most Video LLMs perform comparably to, or even worse than, Image LLMs. This discrepancy arises because existing video benchmarks can often be adequately addressed by single-frame bias~\cite{lei2022revealing,atphard}, without the need for capturing the temporal dynamics of videos. 
To better evaluate the temporal modeling capability, multiple temporal-sensitive benchmarks have been developed~\cite{tempcompass,mvbench,xiao2021nextqa} that cannot be solved by simply relying on single-frame bias as a shortcut.

In this paper, we aim to enhance the temporal modeling ability of Video LLMs and evaluate our model on the temporal-sensitive benchmarks. 
Specifically, we propose a temporal-aware Video LLM (\textbf{TG-Vid}) in this work, featuring a novel \textbf{T}ime \textbf{G}ating module (\textbf{TG}) to enhance temporal modeling. 
This TG module comprises three sub-modules, gating spatial attention, gating temporal attention, and gating MLP, simultaneously capturing spatial and temporal information. 
A recent relevant work ST-LLM~\cite{stllm} also tries to enhance temporal modeling, by directly utilizing BT-Adapter~\cite{liu2024btadapter} which applies spatio-temporal attention in parallel to the vision encoder. 
In contrast, our work builds gating spatio-temporal attention on top of the vision encoder, and our gating mechanism imposes effective module-specific control over each sub-module of the TG module. As validated by experiments, our design achieves better performance on temporal-sensitive benchmarks.

We conduct comprehensive experiments on three temporal-sensitive video benchmarks (\ie, MVBench~\cite{mvbench}, TempCompass~\cite{tempcompass} and NExT-QA~\cite{xiao2021nextqa}). 
The results show that our TG-Vid significantly outperforms the existing Video LLMs across all benchmarks and demonstrate the effectiveness of our TG-Vid on temporal-aware video understanding.
The thorough ablation studies further emphasize that the performance gains are attributed to the designs of our TG module.

\section{Related Work}

\paragraph{Video Large Language Models.}
Benefited from the reasoning power of large language models (LLMs)~\cite{zhang2022opt,2020gpt3,llama,vicuna2023,zhao2023llmsurvey}, Video LLMs~\cite{li2023videochat,maaz2023videochatgpt,videollama,lin2023videollava,tang2023videosurvey, ren2024timechat, wang2024lstp, tan2024koala} have shown impressive performance on video-and-language tasks, such as video question answering~\cite{xu2017MSRVTT_MSVD, jang2017tgif, yu2019activitynet, maaz2023videochatgpt, xiao2021nextqa}.
However, most existing Video LLMs inherit the design of Image LLMs~\cite{zhu2023minigpt4,liu2024llava,InstructBLIP} and overlook the temporal modeling that is critical for video data, leading to unsatisfactory capability on temporal-aware video understanding~\cite{mvbench,tempcompass}. For example, TempCompass~\cite{tempcompass} reveals that the temporal understanding ability of most Video LLMs is on par with or even weaker than Image LLMs. In this work, we propose a temporal-aware Video LLM, featuring a new architecture of time gating module to enhance video temporal modeling.

\smallskip
\paragraph{Video Temporal Modeling.}
Modeling temporal information has been a long-standing topic in video research. Early work utilizes 3D convolutional networks (CNNs) to achieve spatio-temporal video modeling~\cite{carreira2017quo, DBLP:conf/nips/FeichtenhoferPW16, tran2015learning}. To reduce training costs, subsequent CNN-based models explore factorizing convolutions across spatial and temporal dimensions~\cite{sun2015human, tran2019video, tran2018closer, xie2018rethinking, feichtenhofer2020x3d}. 
Further, by leveraging the superiority of Transformer in processing sequences, TimesFormer~\cite{TimesFormer} and ViViT~\cite{vivit} employ Transformer-based architectures to enhance spatio-temporal modeling via spatial and temporal attention.
Beyond single action, a line of work seeks to learn the temporal ordering of actions in procedural activities~\cite{bojanowski2014weakly,chang2020procedure,Zhao2022P3IVPP,zhong2023learning}.
More recently, pretrained image-language models~\cite{2021clip} are transferred to video tasks~\cite{ni2022expanding,pan2022stadapter,luo2022clip4clip,fang2021clip2video,liu2024btadapter}, such as action recognition and video retrieval.
Unlike these works, we extend the idea of spatio-temporal attention to Video LLMs, targeting at temporal-sensitive VideoQA and filling the gap of video modeling in Video LLMs.


\section{Methodology}

In this section, we introduce our \textbf{T}ime \textbf{G}ating \textbf{Vid}eo LLM (\textbf{TG-Vid}).~\cref{fig:overview} provides an overview of our model architecture.
To enhance the temporal modeling of a Video LLM (comprising an LLM, a vision encoder, and a connection module), we propose a \textbf{T}ime \textbf{G}ating (\textbf{TG}) module with a novel module-specific time gating mechanism.

\subsection{Preliminary}

Given a video input with $T$ frames, a pretrained vision encoder~\cite{evaclip} extracts patch embeddings for each frame and concatenates them into video embeddings $\mathbf{V} \in \mathbb{R}^{T \times L_\text{V} \times D_\text{V}}$, where $L_\text{V}$ denotes the number of patch embeddings in each frame and $D_\text{V}$ denotes the dimension of patch embeddings. 
On the other side, given the text input, we employ the text embedder of a pretrained LLM~\cite{vicuna2023} to obtain the text embeddings $\mathbf{T} \in \mathbb{R}^{L_\text{T} \times D_\text{T}}$, where $L_\text{T}$ denotes the number of text tokens and $D_\text{T}$ denotes the dimension of the text embeddings.
This video and text encoding process is common in Video LLMs methods~\cite{mvbench,hawkeye,stllm}.

Our model design extends spatio-temporal attention from ViViT~\cite{vivit} and TimesFormer~\cite{TimesFormer}.
We provide the background knowledge of spatio-temporal attention. 
For clarity, we first formulate the vanilla $N$-layer \textbf{S}patio-\textbf{T}emporal module (\textbf{ST}), which is placed between the vision encoder and the QFormer. 
Each ST layer comprises a \textbf{spatial attention}, a \textbf{temporal attention}, and a two-layer \textbf{MLP}.
Given the input $\mathbf{V}^\ell \in \mathbb{R}^{T \times L_\text{V} \times D_\text{V}}$, 
the $\ell$-th layer of ST ($\mathbf{V}^0$ is set as $\mathbf{V}$) can be formulated as:
\begin{align}
\mathbf{V}^\ell_\text{S} & = \text{ReshapeS}(\mathbf{V}^\ell) \\
\mathbf{Y}^\ell_\text{S} & = \text{MSA}(\text{LN}(\mathbf{V}^\ell_\text{S})) + \mathbf{V}^\ell_\text{S} \\
\mathbf{V}^\ell_\text{T} & = \text{ReshapeT}(\mathbf{Y}^\ell_\text{S}) \\
\mathbf{Y}^\ell_\text{T} & = \text{MSA}(\text{LN}(\mathbf{V}^\ell_\text{T})) + \mathbf{V}^\ell_\text{T} \\
\mathbf{V^\ell_\text{M}} & = \text{ReshapeM}(\mathbf{Y}^\ell_\text{T}) \\
\mathbf{V}^{\ell+1} = \mathbf{Y}^\ell_\text{M} & = \text{MLP}(\text{LN}(\mathbf{V^\ell_\text{M}})) + \mathbf{V^\ell_\text{M}} 
\end{align}
where $\text{LN}(\cdot)$ denotes layer normalization~\cite{ba2016layernorm}, $\text{MSA}(\cdot)$ denotes multi-head self-attention, and $\text{MLP}(\cdot)$ denotes a two-layer MLP. 
$\text{ReshapeS}(\cdot)$ reshapes $\mathbf{V}^\ell \in \mathbb{R}^{T \times L_\text{V} \times D_\text{V}}$ as $\mathbf{V}^\ell_\text{S} \in \mathbb{R}^{T \times L_\text{V} \cdot D_\text{V}}$, 
$\text{ReshapeT}(\cdot)$ reshapes $\mathbf{Y}^\ell_\text{S} \in \mathbb{R}^{T \times L_\text{V} \cdot D_\text{V}}$ as  $\mathbf{V}^\ell_\text{T} \in \mathbb{R}^{L_\text{V} \times T \cdot D_\text{V}}$, 
and $\text{ReshapeM}(\cdot)$ reshapes $\mathbf{Y}^\ell_\text{T} \in \mathbb{R}^{L_\text{V} \times T \cdot D_\text{V}}$ as  $\mathbf{V^\ell_\text{M}} \in \mathbb{R}^{T \times L_\text{V} \times D_\text{V}}$.

\subsection{Time Gating Module (TG)}

The vanilla ST module can model the spatio-temporal information in video inputs. However, directly inserting a randomly initialized ST module into Video LLM results in unstable training and sub-optimal performance.
To address this issue, we propose a novel \textbf{T}ime \textbf{G}ating Module (\textbf{TG}), featuring a \textbf{time gating mechanism} to impose constraints on each sub-module (\ie, a \textbf{gating spatial attention}, a \textbf{gating temporal attention}, and a \textbf{gating MLP}) of the TG module.
These gating sub-modules allow our TG to focus dynamically on relevant information in both spatial and temporal aspects, enhancing the temporal modeling ability of Video LLM.

Unlike previous research works~\cite{sung2022lst,liu2024btadapter} that utilize gating mechanism conditioned solely on a trainable but module-agnostic scalar (\eg, $\alpha \in \mathbb{R}^{1}$) or vector (\eg, $\bm{b} \in \mathbb{R}^{D_\text{V}}$),
the gating function $\text{Gating}(\cdot)$ in our TG is module-specific and conditioned on both the input and output of the sub-module. 
Specifically, \textbf{gating spatial attention} is implemented as:
\begin{equation}
\begin{aligned}
\hat{\mathbf{Y}}^{\ell}_\text{S} &= \text{MSA}(\text{LN}(\mathbf{V}^\ell_\text{S})) \\
\mathbf{Y}^\ell_\text{S} & = \text{Gating}(\mathbf{V}^\ell_\text{S}, \hat{\mathbf{Y}}^{\ell}_\text{S})) + \mathbf{V}^\ell_\text{S} \\
& = \sigma(\text{Cat}(\mathbf{V}^\ell_\text{S}, \hat{\mathbf{Y}}^{\ell}_\text{S})\mathbf{W}_\text{S})  \odot \hat{\mathbf{Y}}^{\ell}_\text{S} + \mathbf{V}^\ell_\text{S}
\end{aligned}
\end{equation}
where $\sigma(\cdot)$ is a sigmoid function, $\text{Cat}(\cdot)$ denotes concatenate operation, $\mathbf{W}_\text{S} \in \mathbb{R}^{2 \cdot D_\text{V} \times D_\text{V}}$ is a linear projection, and $\odot$ denotes element-wise product.
Similarly, \textbf{gating temporal attention} and \textbf{gating MLP} are implemented as follows: 
\begin{equation}
\label{eq:gating_temporal_attention}
\begin{aligned}
\hat{\mathbf{Y}}^{\ell}_\text{T} &= \text{MSA}(\text{LN}(\mathbf{V}^\ell_\text{T})) \\
\mathbf{Y}^\ell_\text{T} 
& = \sigma(\text{Cat}(\mathbf{V}^\ell_\text{T}, \hat{\mathbf{Y}}^{\ell}_\text{T})\mathbf{W}_\text{T})  \odot \hat{\mathbf{Y}}^{\ell}_\text{T} + \mathbf{V}^\ell_\text{T}
\end{aligned}
\end{equation}
\begin{equation}
\begin{aligned}
\hat{\mathbf{Y}}^{\ell}_\text{M} &= \text{MLP}(\text{LN}(\mathbf{V^\ell_\text{M}})) \\
\mathbf{Y}^\ell_\text{M} 
& = \sigma(\text{Cat}(\mathbf{V}^\ell_\text{M}, \hat{\mathbf{Y}}^{\ell}_\text{M})\mathbf{W}_\text{M})  \odot \hat{\mathbf{Y}}^{\ell}_\text{M} + \mathbf{V}^\ell_\text{M}
\end{aligned}
\end{equation}
where $\mathbf{W}_\text{T} \in \mathbb{R}^{2 \cdot D_\text{V} \times D_\text{V}}$ and $\mathbf{W}_\text{M} \in \mathbb{R}^{2 \cdot D_\text{V} \times D_\text{V}}$.

\begin{table*}[t]
    \centering
    \small
    \resizebox{\linewidth}{!}{
        \begin{tabular}{l|cccccccc|cc}
             \toprule
\multicolumn{1}{l|}{\multirow{2}{*}{\textbf{Model}}} & \multicolumn{1}{c}{Otter-V} & \multicolumn{1}{c}{mPLUG-Owl} & \multicolumn{1}{c}{Video-ChatGPT} & \multicolumn{1}{c}{Video-LLaMA} & \multicolumn{1}{c}{VideoChat} & \multicolumn{1}{c}{VideoChat2} & \multicolumn{1}{c}{HawkEye} & \multicolumn{1}{c|}{ST-LLM} & \multirowcell{2}{TG-Vid} & \multirowcell{2}{TG-Vid} \\
&\cite{otter} &\cite{ye2023mplug} & \cite{maaz2023videochatgpt} & \cite{videollama} &\cite{li2023videochat} & \cite{mvbench} & \cite{hawkeye} & \cite{stllm} &&\\
                \bottomrule
\toprule
\textbf{LLM} & LLaMA-7B & LLaMA-7B & Vi-7B & Vi-7B & Vi-7B & Vi-7B & Vi-7B & Vi-7B & Vi-7B & Vi-7B \\
\textbf{\#IT} & - & - & - & - & - & 1.9M & 2.2M  & 220K & 197K & 220K \\
\bottomrule

\toprule
\textbf{Avg} & 26.8 & 29.7 & 32.7 & 34.1 & 35.5 & 51.1 &  47.6 & 54.9 & \underline{56.0} & \textbf{56.4} \\
\bottomrule
        \end{tabular}
    }
    \caption{\textbf{MVBench benchmark experiments.} 
    Comprehensive results are provided in the Appendix~\cref{tab:appendix_MVBench}.
    \#IT denotes instruction tuning samples.
    ``Vi-'' denotes ``Vicuna-''.
    \textbf{Bold}/\underline{underline} denotes the best/second-best result. 
    }
    \label{tab:MVBench}
\end{table*}
\begin{table*}[t]
    \centering
    \small
    \resizebox{\linewidth}{!}{
        \begin{tabular}{l|ccccccccc|cc}
             \toprule
\multicolumn{1}{l|}{\multirow{2}{*}{\textbf{Model}}} & \multicolumn{1}{c}{V-LLaVA} & \multicolumn{1}{c}{LLaMA-VID} & \multicolumn{1}{c}{mPLUG-Owl} & \multicolumn{1}{c}{PandaGPT} & \multicolumn{1}{c}{Valley} & \multicolumn{1}{c}{VideoChat2} & \multicolumn{1}{c}{V-ChatGPT} & \multicolumn{1}{c}{V-LLaMA} & \multicolumn{1}{c|}{ST-LLM$^\clubsuit$} & \multirowcell{2}{TG-Vid} & \multirowcell{2}{TG-Vid} \\
&\cite{lin2023videollava} & \cite{li2023llamavid} & \cite{ye2023mplug} & \cite{su2023pandagpt} & \cite{luo2023valley} & \cite{mvbench} & \cite{maaz2023videochatgpt}& \cite{videollama}& \cite{stllm} &&\\
                             \bottomrule
              \toprule
                \textbf{LLM} & Vi-7B & Vi-7B & LLaMA-7B & Vi-13B & StableVi-7B & Vi-7B & Vi-7B & Vi-13B & Vi-7B & Vi-7B & Vi-7B \\
\textbf{\#IT} & - & - & - & - & - & 1.9M & - & - & 220K & 197K & 220K \\
                \bottomrule
             \toprule
                \textbf{Avg(Caption Matching)} &  63.7 & 53.6 & 49.3 & 51.3 & 22.0 & 55.6 & 51.8 & 53.5 & 64.8 &  \textbf{67.6} & \underline{67.5} \\
                \textbf{Avg(Yes/No QA)} & 56.4 & 53.0 & 54.4 & 51.8 & 53.5 & \underline{58.0} & 50.7 & 53.7  & 54.0 & \textbf{58.1} & 56.8 \\
                \textbf{Avg(Multi-Choice QA)} & 44.7 & 35.3 & 40.0 & 31.1 & 31.8 &  51.1  & 35.2 & 33.9 & \underline{53.7} & {52.9} & \textbf{54.4} \\
             \bottomrule
             \toprule
                \textbf{Avg(ALL)} & 54.9 & 47.3 & 47.9 & 44.7 & 35.8 & 54.9 & 45.9 & 47.0 & {57.5} & \underline{59.5} & \textbf{59.6}\\
             \bottomrule
        \end{tabular}
    }
    \caption{\textbf{TempCompass benchmark experiments.} 
    Comprehensive results are provided in the Appendix~\cref{tab:appendix_TempCompass}.
    \#IT denotes instruction tuning samples.
    ``V-'' denotes ``Video-'' and ``Vi-'' denotes ``Vicuna-''.
    Avg(ALL) is the overall average result, calculated as the average of Avg(Caption Matching), Avg(Yes/No QA), and Avg(Multi-Choice QA). 
    \textbf{Bold}/\underline{underline} denotes the best/second-best average result. $\clubsuit$: We reproduce the training and inference.}
    \label{tab:TempCompass}
\end{table*}

\subsection{Time Gating Video LLM}
By inserting the $N$-layer TG module between the frozen vision encoder and the frozen QFormer, we propose \textbf{TG-Vid}, a \textbf{T}ime \textbf{G}ating \textbf{Vid}eo LLM.
The output video embeddings of the pretrained QFormer are flattened as $\mathbf{V}_\text{Q} \in \mathbb{R}^{T \cdot L_\text{q} \times D_\text{V}}$, where $L_\text{q}$ denotes the length of query tokens for each frame. Subsequently, $\mathbf{V}_\text{Q}$ is projected into the text embedding space and concatenated with the text embedding $\mathbf{T}$ as follows:
\begin{equation}
\mathbf{VT} = [\mathbf{V}_\text{Q}\mathbf{W}_\text{VT}, \mathbf{T}]
\end{equation}
where $\mathbf{W}_\text{VT} \in \mathbb{R}^{D_\text{V} \times D_\text{T}}$ is a trainable linear projection, and $\mathbf{VT}  \in \mathbb{R}^{(T \cdot L_\text{q} + L_\text{T}) \times D_\text{T}}$ is the input into the LLM.
Same as previous Video LLMs, our TG-Vid model is trained on next token prediction.

\section{Experiments}
Compared with existing Video LLMs, we evaluate our TG-Vid on three temporal-sensitive video understanding benchmarks (\ie, MVBench~\cite{mvbench}, TempCompass~\cite{tempcompass}, and NExT-QA~\cite{xiao2021nextqa,atphard}). 
More details of datasets, implementation, experiment results and visualization are provided in the Appendix.

\subsection{Main Results}

Tab.~\ref{tab:MVBench}, Tab.~\ref{tab:TempCompass}, and Tab.~\ref{tab:nextqa_two} show our main results on MVBench, TempCompass, and NExT-QA, respectively. Our TG-Vid model achieves the best performance and surpasses previous methods by a large margin across all benchmarks. 
For example, compared with the closest competitor ST-LLM, our TG-Vid-220K achieves +1.5 on MVBench, +2.1 on TempCompass, +3.2 on NExT-QA ATP-hard, and +3.2 on NExT-QA Val. 
These impressive results demonstrate a consistent finding that our TG-Vid model can capture temporal information more effectively, attributed to the TG designs.

\subsection{Ablation Studies}
Given the comparable performance of TG-Vid-220K and TG-Vid-197K, the ablation studies are based on the latter for efficiency consideration.

\paragraph{TG Module.}
In~\cref{tab:ablate_TG}, row 3 significantly outperforms row 1 by a large margin (+3.0), demonstrating the effectiveness of our TG module in empowering temporal-aware video understanding.

\paragraph{Time Gating Mechanism.}
Row 3 significantly surpasses row 2 (+1.5), underscoring the crucial role of the time gating mechanism in enhancing video temporal modeling.

\paragraph{TG Components.} 
The results in~\cref{tab:ablate_TG} indicate that each sub-module of TG module contributes to performance improvement. 
Notably, the proposed gating temporal attention provides the most significant enhancement (from 54.7 to 56.0), further validating the necessity of temporal modeling.

\begin{table}[t]
\centering
\resizebox{\columnwidth}{!}{
    \begin{tabular}{l|c|ccc|cccc}
    \toprule
\multicolumn{1}{l|}{\multirow{2}{*}{\textbf{Model}}} & \multirowcell{2}{\textbf{\#IT}} & \multicolumn{3}{c}{\textbf{NExT-QA ATP-hard}} & \multicolumn{4}{c}{\textbf{NExT-QA Val}} \\
     & & Acc@\textbf{C} & Acc@\textbf{T} & Acc@\textbf{All} & Acc@\textbf{C} & Acc@\textbf{T} & Acc@\textbf{D} & Acc@\textbf{All}  \\
\bottomrule
\toprule
VFC$^\spadesuit$~\cite{yang2021just} & -  & 32.2 & 30.0 & 31.4 & 49.6 & 51.5 & 63.2 & 52.3 \\
ATP~\cite{atphard} & -  & 38.4 & 36.5 & 38.8  & 53.1 & 50.2 & 66.8 & 54.3 \\
GF~\cite{bai2024glance} & - & 48.7 & 50.3 & 49.3  & 56.9 & 57.1 & 70.5 & 58.8 \\
SeViT~\cite{kim2023semi} & -  & 43.3 & 46.5 & -  & 54.0 & 54.1 & 71.3 & 56.7 \\
HiTeA~\cite{ye2023hitea} & -  & 47.8 & 48.6 & -  & 62.4 & 58.3 & 75.6 & 63.1 \\
VideoAgent$^\spadesuit$~\cite{wang2024videoagent} & -  & 57.8 & 58.8 & 58.4  & 72.7 & 64.5 & 81.1 & 71.3 \\
SEVILA~\cite{yu2024self} & -  & - & - & - & 74.2 & 69.4 & \underline{81.3} & 73.8 \\
VideoChat2~\cite{mvbench} & 1.9M  & - & - & - & 68.7 & 64.7 & 76.1 & 68.6 \\
HawkEye~\cite{hawkeye} & 2.2M  & - & - & - & - & - & - & 67.9 \\
ST-LLM$^\clubsuit$~\cite{stllm} & 220k  & 65.5 & 61.9 & 64.0 & 74.3 & 70.0 & \underline{81.3} & 74.0 \\
\bottomrule
\toprule
TG-Vid & 197k  & \underline{68.4} & \textbf{66.3} & \textbf{67.5} & \textbf{77.4} & \textbf{73.8} & \textbf{84.3} & \textbf{77.3} \\
TG-Vid & 220K  & \textbf{68.5} & \underline{65.2} & \underline{67.2} & \underline{77.3} & \underline{73.5} & \textbf{84.3} & \underline{77.2} \\
\bottomrule
    \end{tabular}
    }
    \caption{\textbf{Experiments on NExT-QA ATP-hard subset and NExT-QA validation dataset.} \textbf{C}, \textbf{T}, and \textbf{D} are causal, temporal, and descriptive subsets, respectively. \textbf{Bold}/\underline{underline} denotes the best/second-best result.  $\clubsuit$: We reproduce the training and inference. $\spadesuit$: Zero-shot.}
    \label{tab:nextqa_two}
\end{table}

\begin{table}[t]
\centering
\resizebox{0.99\columnwidth}{!}{
    \begin{tabular}{ccc|c|c|c}
    \toprule
    \multicolumn{3}{c|}{\textbf{TG Components}} & \multicolumn{1}{c|}{\textbf{Gating}} & \multirowcell{2}{\textbf{\#IT}}   & \multicolumn{1}{c}{\textbf{MVBench}}  \\  
      \textbf{Spatial} & \textbf{Temporal} & \textbf{MLP} & \textbf{Mechanism} & & \textbf{Avg} \\
          \bottomrule
\toprule
\xmark & \xmark &\xmark &\xmark& 197K  & 53.0  \\ 
\cmark& \cmark & \cmark &\xmark & 197K  & {54.5}    \\
\cmark& \cmark & \cmark &\cmark & 197K  & \textbf{56.0}   \\
\toprule
\xmark & \cmark &\cmark &\cmark & 197K  & 55.6 \\
\cmark & \xmark & \cmark &\cmark & 197K  & 54.7 \\
\cmark & \cmark & \xmark &\cmark & 197K  &  {55.7}  \\
    \bottomrule

    \end{tabular}
    }
    \caption{\textbf{Ablation studies on TG module}. 
    }
    \label{tab:ablate_TG}
\end{table}

\section{Conclusion}
In this paper, we focus on developing a Video LLM, \textbf{TG-Vid}, to overcome the struggles of the existing Video LLMs in temporal-aware video understanding.
Specifically, we propose a novel \textbf{T}ime \textbf{G}ating module (\textbf{TG}) with a time gating mechanism, to enhance the temporal modeling ability of TG-Vid.
Comprehensive experiments and ablation studies conducted on three temporal-sensitive benchmarks (\ie, MVBench, TempCompas, and NExT-QA) indicate that TG-Vid outperforms the existing Video LLMs by a large margin.
These results demonstrate the effectiveness of our TG design in enhancing temporal modeling, thereby empowering our TG-Vid with a strong ability of temporal-aware video understanding.


\smallskip 
\noindent\textbf{Limitations.} 
Our proposed TG-Vid model has achieved strong performance on the temporal-sensitive video understanding benchmarks. However, there are still some limitations: 
(1) Despite that our TG module can significantly enhance the temporal modeling of the Video LLM, integrating it into Video LLM requires additional computation; 
(2) Similar to the existing Video LLMs, our TG-Vid model has the potential to inherit the undesired biases from the training dataset and the pretrained LLMs; 
(3) The focus of this work is on temporal modeling. Whether the proposed TG-Vid model and the TG module can be generalized to other video-and-language tasks, such as long video understanding, is worth exploring in future research.

\smallskip 
\noindent\textbf{Acknowledgements.} 
This work was supported by National Key R\&D Program of China (Project No. 2022ZD0161200, 2022ZD0161201). This work is also supported by Hong Kong Research Grant Council - Early Career Scheme (Grant No. 24200223) and Hong Kong Innovation and Technology Commission Project No. ITS/228/22FP. This work was also partially funded by the Centre for Perceptual and Interactive Intelligence (CPII) Ltd under the Innovation and Technology Commission (ITC)’s InnoHK. Prof. Liwei Wang is also a PI of CPII.
This work was also partially supported by the Research Grants Council of the Hong Kong Special Administrative Region, China (No. CUHK 14206921 of the General Research Fund).

\bibliography{egbib}

\clearpage

\appendix

\smallskip
\noindent {\LARGE {{\rm {\bf Appendix}}}}

\bigskip

In the appendix, we provide more details of:
(1) the statistics of the training dataset; 
(2) the implementation details and hyper-parameters for training; 
(3) additional ablation study on the number of layers of the proposed TG module; 
(4) additional ablation study on the training output format; (5) additional ablation study on the designs of TG module; 
(6) the comprehensive results of MVBench (comprising twenty sub-tasks) and TempCompass (comprising three sub-tasks);
(7) the visualization of time gating.

\section{Training Dataset Statistics} \label{sec:appendix_train_data}
To ensure a fair comparison with the state-of-the-art Video LLM, ST-LLM~\cite{stllm}, our TG-Vid-220K model utilizes the same training dataset as ST-LLM, as detailed in~\cref{tab:training_data}.
For the training dataset of our TG-Vid-197K model, we filter out the Conversation-VideoChatGPT and VQA-WebVidQA datasets to improve training efficiency.

Moreover, the training dataset with 220K video-text pairs is also a subset of the training dataset of VideoChat2~\cite{mvbench}, which contains 1.9M video-text pairs.

\section{Implementation Details} \label{sec:implementation}
Following ST-LLM, we adopt the Vicuna-7B-v1.1~\cite{vicuna2023} as our pretrained LLM and the EVA-ViT-g~\cite{evaclip} as our pretrained vision encoder. 
The QFormer is also initialized from the pretrained InstructBLIP~\cite{InstructBLIP}, while our TG module is randomly initialized. 
Following the designs of LLaMA~\cite{llama}, the self-attention inside the TG module is implemented as self-attention with rotary position embeddings (RoPE)~\cite{rope}.
Similarly, the activation function of MLP inside the TG module is implemented as the SwiGLU activation function.
Our TG-Vid model is subsequently trained on the video instruction tuning dataset, which is described in~\cref{sec:appendix_train_data}.

During training, the vision encoder and the QFormer are frozen, while other modules of TG-Vid are trainable.
The training of TG-Vid-197K costs about 7 hours and the training of TG-Vid-220K costs about 13 hours.
Both of these trainings are conducted on 8 A100 GPUs (each GPU has 80G memory).

Moreover, TG-Vid-197K and TG-Vid-220K share the same hyper-parameters for training, which are listed in~\cref{tab:training_param}.

\begin{table}[t]
\centering
\resizebox{\columnwidth}{!}{
    \begin{tabular}{l|l|c}
    \toprule
\textbf{Category} & \textbf{Training Dataset} & \textbf{\#Video-Text Pairs} \\
\bottomrule
\toprule
Conversation & VideoChatGPT$^\diamondsuit$ & 13,303 \\
Classification & Kinetics-710 & 40,000 \\
Classification & SthSthV2 & 40,000 \\
Reasoning & NExTQA & 34,132 \\
Reasoning & CLEVRER\_QA & 40,000 \\
Reasoning & CLEVRER\_MC & 42,620 \\
VQA & WebVidQA$^\diamondsuit$ & 10,000 \\
\bottomrule
\toprule
Total & - & 220055 \\
\bottomrule
    \end{tabular}
    }
    \caption{The statistics of the training dataset with 220K video-text pairs. $\diamondsuit$ denotes the datasets that are filtered out in the training dataset with 197K video-text pairs.}
    \label{tab:training_data}
\end{table}

\begin{figure}[t]
\begin{center}
  \includegraphics[width=0.9\columnwidth]{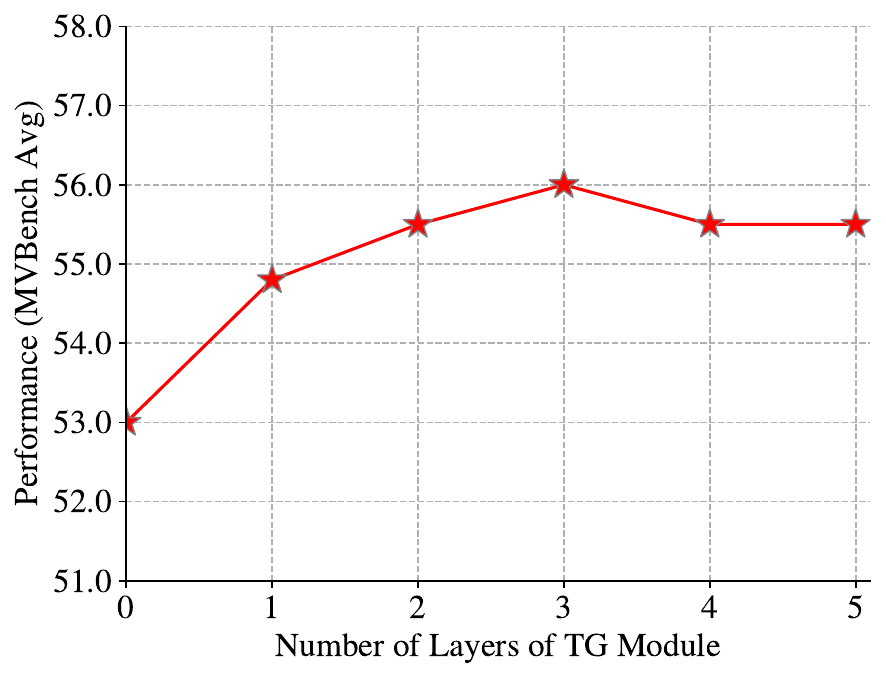}
\end{center}
   \caption{Ablation study on the number of layers of the TG module based on TG-Vid-197K.}
\label{fig:ablate}
\end{figure}

\begin{table}[t]
\centering
\resizebox{0.9\columnwidth}{!}{
    \begin{tabular}{l|c}
    \toprule
\textbf{Hyper-parameter} & \textbf{TG-Vid Training} \\
\bottomrule
\toprule
Number of Layers of TG & $N=3$ \\
Training Epochs & 2 \\
Btach Size & 64 \\
Input Frame & 16 \\
Input Resolution & 224 $\times$ 224 \\
Max Text Length & 256 \\
Max Model Length & 1024 \\
Optimizer & AdamW \\
Optimizer Momentum & $\beta_1, \beta_2{=}0.9, 0.999$ \\
Weight Decay & 0.0 \\
Learning Rate Schedule & Cosine \\
Learning Rate & 2e-5 \\
Warmup Ratio & 0.03 \\
QFormer Query Token Length & $L_q = 32$\\
\bottomrule
    \end{tabular}
    }
    \caption{Hyper-parameters for TG-Vid training.}
    \label{tab:training_param}
\end{table}
\begin{table}[t]
\centering
\resizebox{\columnwidth}{!}{
    \begin{tabular}{l|c|c}
    \toprule
\textbf{Model} & \textbf{Training Output Format} & \textbf{MVBench Avg} \\
\bottomrule
\toprule
TG-Vid-197K & ``(A) chase the dog.'' & 56.3 \\
TG-Vid-197K & ``A'' & 56.0 \\
\bottomrule
    \end{tabular}
    }
    \caption{Experiments on different training output formats.}
    \label{tab:output_format}
\end{table}
\begin{table}[!h]
\centering
\resizebox{\columnwidth}{!}{
    \begin{tabular}{l|c|c|c}
    \toprule
\textbf{Model} & \textbf{Self-Attention with RoPE} & \textbf{MLP with SwiGLU} & \textbf{MVBench Avg} \\
\bottomrule
\toprule
TG-Vid-197K & \xmark & \cmark & 56.3 \\
TG-Vid-197K & \cmark & \xmark & 56.0 \\
\bottomrule
\toprule
TG-Vid-197K & \cmark & \cmark & 56.0 \\
\bottomrule
    \end{tabular}
    }
    \caption{Experiments on the designs borrowed from LLMs.}
    \label{tab:llm_design}
\end{table}

\section{Number of Layers of TG} 
As shown in~\cref{fig:ablate}, we ablate the depth of the TG module. The results reveal that all of the models with the TG module significantly surpass the model without the TG module, demonstrating the effectiveness of our proposed TG module in empowering temporal-aware video understanding.
Moreover, the results also indicate that the 3-layer TG module achieve the best performance. Therefore, we use $N = 3$ in~\cref{tab:training_param} by default.

\section{Training Output Format} 
To improve the training efficiency for LLM decoding, we explore modifying the training output format.
The original output format form VideoChat2-IT~\cite{mvbench} is ``(A) chase the dog.'', while 
ours is modified as a direct output format: ``A''. 
Following the instruction used in LLaVA1.5~\cite{liu2024improved}, we also add an instruction ``Answer with the option’s letter from the given choices directly.'' in the text input.

As shown in~\cref{tab:output_format}, the performance after modification is slightly decreased but the performance is still comparable.
Therefore, we utilize the direct output format as our training output format for efficiency consideration.

\section{Designs Borrowed from LLaMA} 
As mentioned in~\cref{sec:implementation}, we borrow two designs (\ie, RoPE and SwiGLU) from LLaMA into the implementation of the TG module.
In this section, we ablate these designs, as shown in~\cref{tab:llm_design}. 
The results demonstrate that these designs can slightly improve the model performance. 
Therefore, we introduce these designs into the implementation of our TG module.

\section{Comprehensive Results of MVBench and TempCompass.} 
Due to the space limitation of the main paper, we present the comprehensive results of MVBench (comprising twenty sub-tasks) and TempCompass in (comprising three sub-tasks) in~\cref{tab:appendix_MVBench} and~\cref{tab:appendix_TempCompass}, respectively.

\begin{table*}[t]
    \centering
    \small
    \resizebox{\linewidth}{!}{
        \begin{tabular}{l|cccccccc|cc}
             \toprule
\textbf{Model} & Otter-V & mPLUG-Owl & Video-ChatGPT & Video-LLaMA & VideoChat & VideoChat2 & HawkEye & ST-LLM & TG-Vid & TG-Vid \\
                \bottomrule
\toprule
\textbf{LLM} & LLaMA-7B & LLaMA-7B & Vi-7B & Vi-7B & Vi-7B & Vi-7B & Vi-7B & Vi-7B & Vi-7B & Vi-7B \\
\textbf{\#IT} & - & - & - & - & - & 1.9M & 2.2M  & 220K & 197K & 220K \\
\bottomrule
             \toprule
                \textbf{Action} &&&&&&&& \\
Action Sequence & 23.0 & 22.0 & 23.5 & 27.5 & 33.5 & 66.0 & - & 66.0 & \textbf{72.5} & \underline{70.5} \\
Action Prediction & 23.0 & 28.0 & 26.0 & 25.5 & 26.5 & 47.5 & - & 53.5 & \textbf{57.0} & \underline{54.5}\\
Action Antonym & 27.5 & 34.0 & 62.0 & 51.0 & 56.0 & 83.5 & - &84.0 & \underline{85.0} & \textbf{87.5} \\
Fine-grained Action & 27.0 & 29.0 & 22.5 & 29.0 & 33.5 & \textbf{49.5} & - & 44.0 & 45.0 & \underline{46.0} \\
Unexpected Action & 29.5 & 29.0 & 26.5 & 39.0 & 40.5 & \textbf{60.0} & - &  \underline{58.5} & 53.5 & 57.5\\
\bottomrule
             \toprule
                \textbf{Object} &&&&&&&& \\
Object Existence & 53.0 & 40.5 & 54.0 & 48.0 & 53.0 & 58.0 & - & 80.5 & \textbf{83.5} & \underline{83.0}\\
Object Interaction & 28.0 & 27.0 & 28.0 & 40.5 & 40.5 & 71.5 & - & {73.5} & \textbf{74.5} & \underline{74.0}\\
Object Shuffle & 33.0 & 31.5 & \underline{40.0} & 38.0 & 30.0 & \textbf{42.5} & - & 38.5 & 36.0 & 36.5\\
\bottomrule
             \toprule
                \textbf{Position} &&&&&&&& \\
Moving Direction & 24.5 & 27.0 & 23.0 & 22.5 & 25.5 & 23.0 & - & {42.5} & \textbf{45.5} & \underline{45.0} \\
Action Localization & 23.5 & 23.0 & 20.0 & 22.5 & {27.0} & 23.0 & - & \underline{31.0} & \textbf{32.0} &  {29.5} \\
\bottomrule
             \toprule
                \textbf{Scene} &&&&&&&& \\
Scene Transition & 27.5 & 29.0 & 31.0 & 43.0 & 48.5 & \textbf{88.5}  & - & \underline{86.5} & 82.0 & {85.5}\\
\bottomrule
             \toprule
                \textbf{Count} &&&&&&&& \\
Action Count & 26.0 & 31.5 & 30.5 & 34.0 & 35.0 & \textbf{39.0} & - & \underline{36.5} & 32.5 & 36.0\\
Moving Count & 28.5 & 27.0 & 25.5 & 22.5 & 20.5 & 42.0 & - & {56.5} & \textbf{68.5} & \underline{66.5}\\
\bottomrule
             \toprule
                \textbf{Attribute} &&&&&&&& \\
Moving Attribute & 18.0 & 40.0 & 39.5 & 32.5 & 42.5 & 58.5 & - & {78.5} & \underline{82.0} & \textbf{85.0}\\
State Change & 38.5 & 44.0 & \textbf{48.5} & 45.5 & {46.0} & 44.0 & - & \underline{43.0} & 40.5 & 46.0 \\
\bottomrule
             \toprule
                \textbf{Pose} &&&&&&&& \\
Fine-grained Pose & 22.0 & 24.0 & 29.0 & 32.5 & 26.5 & \textbf{49.0} & - &  44.5 & \underline{47.5} & 42.0\\
\bottomrule
             \toprule
                \textbf{Character} &&&&&&&& \\
Character Order & 22.0 & 31.0 & 33.0 & 40.0 & 41.0 & 36.5 &  - & {46.5} & \textbf{47.5} & \underline{47.0} \\
\bottomrule
             \toprule
                \textbf{Cognition} &&&&&&&& \\
Egocentric Navigation & 23.5 & 26.0 & 29.5 & 30.0 & 23.5 & \underline{35.0} &  - & {34.5} &{33.0} & \textbf{37.0} \\
Episodic Reasoning & 19.0 & 20.5 & 26.0 & 21.0 & 23.5 & \underline{40.5} &  - & \textbf{41.5} & \textbf{41.5} & 40.0 \\
Counterfactual Inference & 19.5 & 29.5 & 35.5 & 37.0 & 36.0 & \textbf{65.5} & - &  58.5 & \underline{60.0} & 58.0\\
\bottomrule
\toprule
\textbf{Avg} & 26.8 & 29.7 & 32.7 & 34.1 & 35.5 & 51.1 &  47.6 & 54.9 & \underline{56.0} & \textbf{56.4} \\
\bottomrule
        \end{tabular}
    }
    \caption{\textbf{Comprehensive results on the MVBench benchmark.} Experiments are conducted on 20 MVBench sub-tasks.
    \#IT denotes instruction tuning samples.
    ``V-'' in the Model names denotes ``Video-'' and ``Vi-'' in the LLM names denotes ``Vicuna-''.
    Bold/underline denotes the best/second-best result.
    }
    \label{tab:appendix_MVBench}
\end{table*}
\begin{table*}[t]
    \centering
    \small
    \resizebox{\linewidth}{!}{
        \begin{tabular}{l|ccccccccc|cc}
             \toprule
             \textbf{Model} & V-LLaVA & LLaMA-VID & mPLUG-Owl & PandaGPT & Valley & VideoChat2 & V-ChatGPT & V-LLaMA & ST-LLM$^\clubsuit$ & TG-Vid & TG-Vid \\
                             \bottomrule
              \toprule
                \textbf{LLM} & Vi-7B & Vi-7B & LLaMA-7B & Vi-13B & StableVi-7B & Vi-7B & Vi-7B & Vi-13B & Vi-7B & Vi-7B & Vi-7B \\
\textbf{\#IT} & - & - & - & - & - & 1.9M & - & - & 220K & 197K & 220K \\
                \bottomrule
             \toprule
                \textbf{Caption Matching} &&&&&&&&&& \\
             \midrule
                Action  & 88.2 & 72.7 & 56.9 & 56.6 & 15.5 & 65.0 & 64.6 & 73.1 & 93.9 & 96.0 & 95.3 \\
                Direction  & 53.8 & 45.6 & 45.3 & 51.4 & 21.4 & 53.8 & 48.6 & 47.4 & 59.3 & 54.7 &55.4 \\
                Speed &  61.9  & 52.2 & 46.4 & 44.3 & 22.0 & 52.6 & 47.8 & 47.1 & 54.3 & 57.0 &58.1  \\
                Event Order  &  57.0  & 49.0 & 49.3 & 55.0 & 28.3 & 53.0 & 49.3 & 52.0 & 55.0 & 65.3 &62.0 \\
                Attribute Change  &  58.3  & 49.0 & 49.0 & 49.0 & 22.9 & 53.8 & 48.6 & 48.3 & 61.5 &  64.9 &66.7\\
                \midrule
                Avg &  63.7 & 53.6 & 49.3 & 51.3 & 22.0 & 55.6 & 51.8 & 53.5 & 64.8 &  \textbf{67.6} & \underline{67.5} \\
             \bottomrule
             \toprule
                \textbf{Yes/No QA} &&&&&&&&&& \\
             \midrule
                Action  & 74.3 & 63.0 & 64.4 & 53.0 & 58.1 & 72.8 & 52.5 & 68.1 & 68.1 & 77.4 & 76.5 \\
                Direction  & 51.8 & 48.8 & 50.6 & 49.6 & 52.0 & 53.8 & 50.0 & 46.0 & 50.6 & 51.6 & 50.6 \\
                Speed & 50.3 & 49.2 & 51.2 & 50.8 & 52.5 & 53.8 & 49.5 & 48.8 & 49.9 & 52.5 &52.7 \\
                Event Order  & 49.2 & 48.4 & 51.3 & 53.7 & 50.3 & 51.3 & 51.0 & 51.8 & 50.0 & 55.0 & 50.0 \\
                Attribute Change  & 51.1 & 52.7 & 52.0 & 52.2 & 52.9 & 53.8 & 50.0 & 50.9 & 51.6 & 54.0 & 54.0 \\
                \midrule
                Avg & 56.4 & 53.0 & 54.4 & 51.8 & 53.5 & \underline{58.0} & 50.7 & 53.7  & 54.0 & \textbf{58.1} & 56.8 \\
             \bottomrule
             \toprule
                \textbf{Multi-Choice QA} &&&&&&&&&& \\
             \midrule
                Action  & 70.4 & 58.6 & 66.6 & 35.5 & 47.0 & 88.5 & 47.0 & 54.1 & 92.0 & 92.6 & 91.1\\
                Direction  & 32.2 & 29.9 & 29.3 & 27.8 & 29.3 & 36.4 & 31.6 & 24.5 & 37.3 & 34.9 & 39.4\\
                Speed  & 38.2 & 29.3 & 32.2 & 29.3 & 32.5 & 42.0 & 28.4 & 28.1 & 46.7 & 45.4 & 46.4\\
                Event Order &  41.4  & 30.5 & 34.8 & 31.8 & 18.9 &  40.7  & 37.1 & 32.8 & 42.7 & 41.1 & 43.0\\
                Attribute Change  & 39.9 & 26.0 & 35.4 & 30.9 & 29.9 &  45.5  & 30.9 & 28.5  & 50.0 &  50.3 & 52.1 \\
             \midrule
                Avg & 44.7 & 35.3 & 40.0 & 31.1 & 31.8 &  51.1  & 35.2 & 33.9 & \underline{53.7} & {52.9} & \textbf{54.4} \\
             \bottomrule
             \toprule
                \textbf{Avg(ALL)} & 54.9 & 47.3 & 47.9 & 44.7 & 35.8 & 54.9 & 45.9 & 47.0 & {57.5} & \underline{59.5} & \textbf{59.6}\\
             \bottomrule
        \end{tabular}
    }
    \caption{\textbf{Comprehensive results on the TempCompass benchmark.} Experiments are conducted on three TempCompass tasks: Caption Matching, Yes/No QA,  and Multi-Choice QA.
    \#IT denotes instruction tuning samples.
    ``V-'' in the Model names denotes ``Video-'' and ``Vi-'' in the LLM names denotes ``Vicuna-''.
    The overall average result \textbf{Avg(ALL)} is calculated as the average of Avg(Caption Matching), Avg(Yes/No QA), and Avg(Multi-Choice QA). Bold/underline denotes the best/second-best result. $\clubsuit$: We reproduce the training and inference.}
    \label{tab:appendix_TempCompass}
\end{table*}

\section{Visualization of Time Gating.} 
To gain more insight into how time gating works, we provide some visualizations in xx. 
To be specific, we visualize the heatmap of the gate values produced by our  time gating mechanism, based on the following steps:

\begin{itemize}
    \item Step 1: Randomly select an input sample (with a $T$-frame video).
    \item Step 2: For the input sample, obtain its gate values of the gating temporal attention in the first layer of TG (\ie, the result of the sigmoid function in \Cref{eq:gating_temporal_attention}), denoted as $\mathbf{G} \in \mathbb{R}^{L_\text{V} \times T \cdot D_\text{V}}$. $L_\text{V}$ denotes the number of patch embeddings in each frame and $D_\text{V}$ denotes the dimension of patch embeddings.
    \item Step 3: Visualize the gate values. Specifically, reshape $\mathbf{G}$ as $\mathbb{R}^{L_\text{V} \times T \times D_\text{V}}$ and adopt average-pooling along the $D_\text{V}$ dimension to obtain $\hat{\mathbf{G}} \in \mathbb{R}^{L_\text{V} \times T}$. Finally, visualize $\hat{\mathbf{G}}$ with a heatmap visualization $H$.
\end{itemize}

To conduct a detailed analysis, we repeated steps 1-3 twice. Given two different input samples from MVBench, SampleA (with a $T_A$-frame video) and SampleB (with a $T_B$-frame video), we obtain $\hat{\mathbf{G}}_\text{A}$ and $\hat{\mathbf{G}}_\text{B}$, and visualize the corresponding heatmaps $H_A$ and $H_B$. Our observations are as follows:
\begin{itemize}
    \item For videos with different durations (\ie, different numbers of frames), the shape of $\hat{\mathbf{G}}$ can adapt accordingly ($\hat{\mathbf{G}}_\text{A} \in \mathbb{R}^{L_\text{V} \times T_A}$ and $\hat{\mathbf{G}}_\text{B} \in \mathbb{R}^{L_\text{V} \times T_B}$). Therefore, for different input samples, {the time gating mechanism can adapt to the content of specific input and performs temporal-sensitive control, thereby enhancing the temporal modeling ability of the model}.
    
    \item For all patch embeddings at the same time (\ie, the same frame), the corresponding gate values of $\hat{\mathbf{G}}$ change dynamically, revealing that {the time gating mechanism can discern the information at different spatial locations and provides dynamic, fine-grained control}.
    
    \item For all patch embeddings at the same spatial location, the corresponding gate values of $\hat{\mathbf{G}}$ also change dynamically, which demonstrates that {the time gating mechanism can also distinguish the temporal dynamics and provide fine-grained control}.
\end{itemize}

\begin{figure*}[htbp]
    \centering
    \begin{subfigure}{0.49\textwidth}
        \centering
        \includegraphics[width=\textwidth]{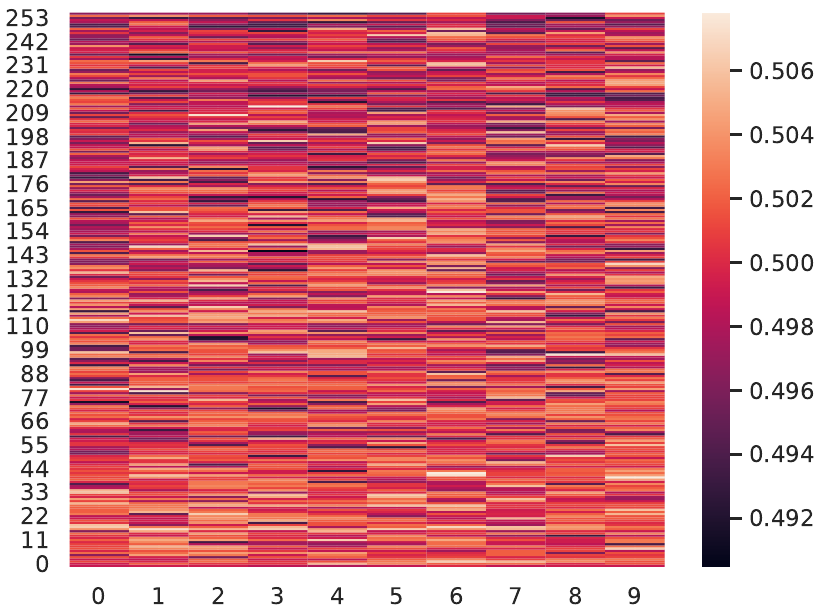} 
        \caption{Input sample with a 10-frame video.}
        \label{fig:sample_a}
    \end{subfigure}
    \hfill
    \begin{subfigure}{0.49\textwidth}
        \centering
        \includegraphics[width=\textwidth]{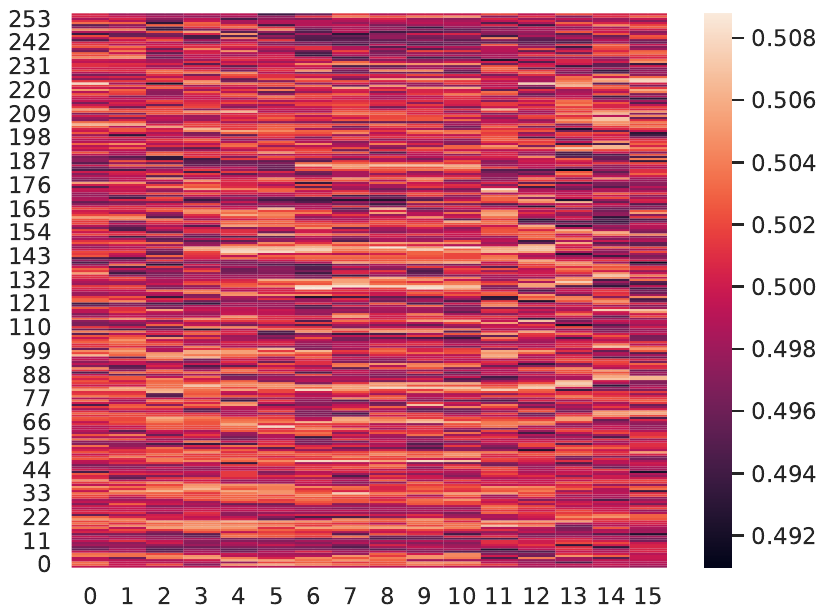} 
        \caption{Input sample with a 16-frame video.}
        \label{fig:sample_b}
    \end{subfigure}
    \caption{Visualization of Time Gating.}
    \label{fig:two_samples}
\end{figure*}

\end{document}